\documentclass[iicol,sn-mathphys]{sn-jnl}

\usepackage{graphicx}%
\usepackage{multirow}%
\usepackage{amsmath,amssymb,amsfonts}%
\usepackage{amsthm}%
\usepackage{mathrsfs}%
\usepackage[title]{appendix}%
\usepackage{xcolor}%
\usepackage{textcomp}%
\usepackage{manyfoot}%
\usepackage{booktabs}%
\usepackage{algorithm}%
\usepackage{algorithmicx}%
\usepackage{algpseudocode}%
\usepackage{listings}%
\usepackage{booktabs}
\usepackage{wrapfig}
\usepackage{caption}
\floatname{algorithm}{Algorithm}

\floatname{algorithm}{Algorithm}

\theoremstyle{thmstyleone}%
\theoremstyle{thmstyletwo}%

\theoremstyle{thmstylethree}%
\usepackage[utf8]{inputenc} 
\usepackage[T1]{fontenc}    
\usepackage{hyperref}       
\usepackage{url}            
\usepackage{booktabs}       
\usepackage{amsfonts} 
\begin{document}

\title[Continual Face Forgery Detection via Historical Distribution Preserving]{Continual Face Forgery Detection via Historical Distribution Preserving}


\author[1]{\fnm{Ke} \sur{Sun}}

\author[2]{\fnm{Shen} \sur{Chen}}
\author[2]{\fnm{Taiping} \sur{Yao}}
\author[1]{\fnm{Xiaoshuai} \sur{Sun}}
\author[2]{\fnm{Shouhong} \sur{Ding}}
\author[*1]{\fnm{Rongrong} \sur{Ji}}

\affil*[1]{Key Laboratory of Multimedia Trusted Perception and Efficient Computing, \\
   \normalsize Ministry of Education of China, Xiamen University, China}

\affil[2]{Youtu Lab, Tencent, China}



\abstract{Face forgery techniques have advanced rapidly and pose serious security threats. Existing face forgery detection methods try to learn generalizable features, but they still fall short of practical application. Additionally, finetuning these methods on historical training data is resource-intensive in terms of time and storage.
In this paper, we focus on a novel and challenging problem: Continual Face Forgery Detection (CFFD), which aims to efficiently learn from new forgery attacks without forgetting previous ones. Specifically, we propose a Historical Distribution Preserving (HDP) framework that reserves and preserves the distributions of historical faces. To achieve this, we use universal adversarial perturbation (UAP) to simulate historical forgery distribution, and knowledge distillation to maintain the distribution variation of real faces across different models. We also construct a new benchmark for CFFD with three evaluation protocols. Our extensive experiments on the benchmarks show that our method outperforms the state-of-the-art competitors.}





\maketitle

\section{Introduction}\label{sec1}


Over the past decades, face forgery methods have made significant strides, capturing the interest of both the academic and industrial realms~\cite{thies2015real,rossler2019faceforensics++,dolhansky2020deepfake}. These techniques have the prowess to create ultra-realistic forged faces, so convincing at times that they can easily deceive the human eye. This verisimilitude has far-reaching implications, giving rise to potential malicious misuse. Whether used in privacy infringements, identity fraud, or other deceptive practices, they pose severe societal challenges. Therefore, the need to engineer potent methods capable of differentiating real faces from their forged counterparts has become increasingly paramount.
Recently, many significant 
face forgery detection methods have been proposed to mine the subtle artifacts~\cite{chen2021local,qian2020thinking,dolhansky2020deepfake,dang2020detection,afchar2018mesonet,sun2021domain,luo2023beyond,luo2021capsule} and
achieved extraordinary performance under known forgery types. However, they all suffer from significant performance degradation when testing under new forgery attacks. Some methods have attempted to learn generalized representation~\cite{li2020face,sun2021dual,luo2021generalizing,sun2021domain,sun2023towards}, 
but their performance on unknown attacks is still far from practical application.
As face forgery techniques continue to evolve, acquiring a comprehensive dataset of all forgery methods and their respective manipulation techniques becomes challenging, especially since such data is often accrued over time. Given this scenario, continually updating models to integrate both past and current data can be both time-consuming and storage-intensive. Relying solely on recent forgery data for model updates introduces the risk of catastrophic forgetting—a situation where the model forgets its previously learned patterns. These limitations present significant barriers to the practical deployment of face forgery detection systems in real-world scenarios.



\begin{figure*}[t!]
    \begin{center}
       \includegraphics[width=0.9\textwidth]{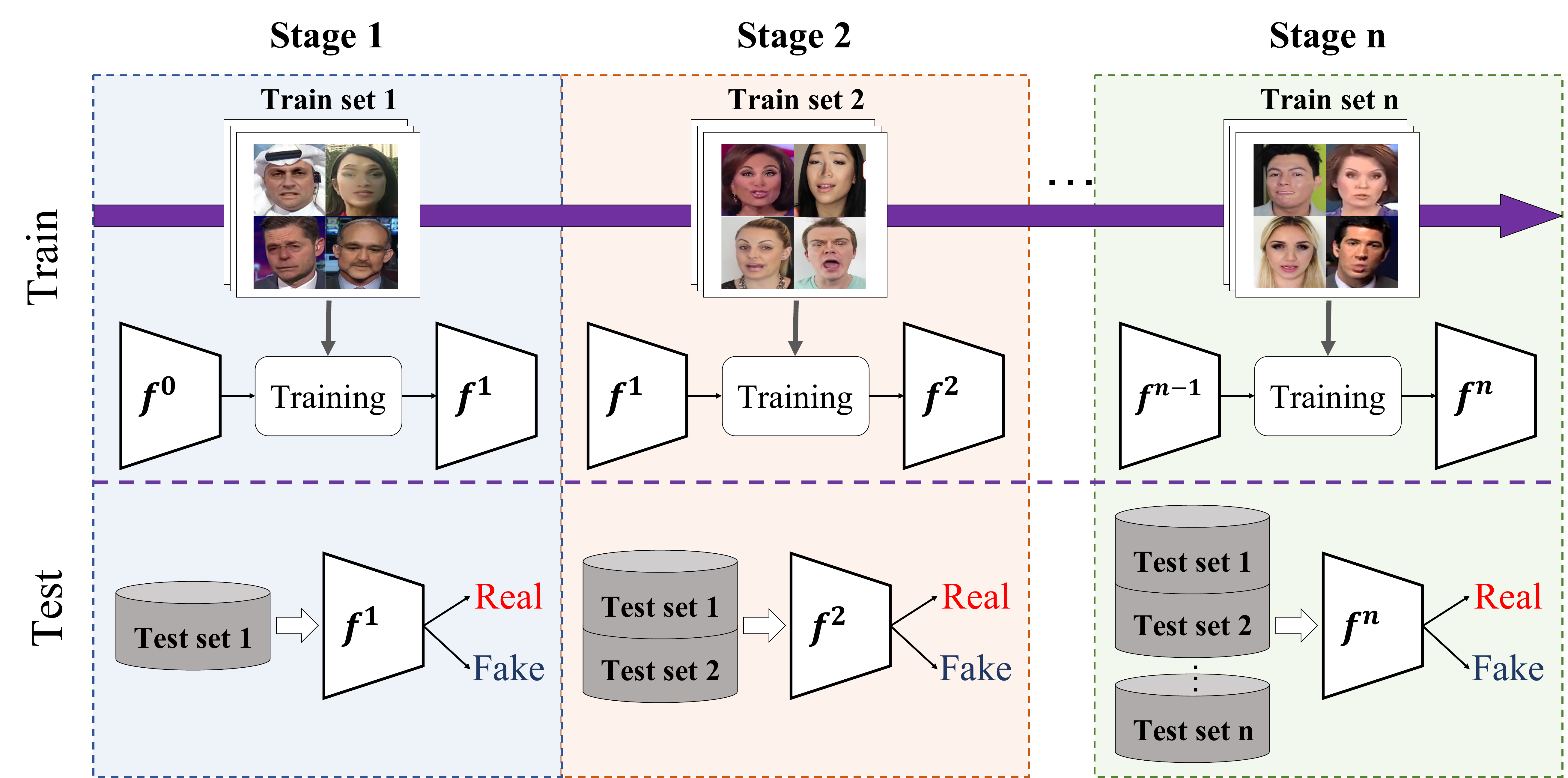}
    \end{center}
       \caption{Pipeline of the training and testing processes in the Continual Face Forgery Detection (CFFD) setting.
       (Best viewed in color.)
       }
    \label{fig:intro}
\end{figure*}


To address this challenge, we introduce and tackle a novel and pressing problem: \textbf{Continual Face Forgery Detection} (CFFD)~\cite{li2023continual}. CFFD evaluates the capacity of detectors to adapt to new attack techniques sequentially while retaining proficiency in recognizing earlier ones. As depicted in Fig.\ref{fig:intro}, diverse training data are introduced across different stages. The goal of the model is to assimilate new information without discarding what it has previously learned.
While certain existing continual learning methods might seem applicable to CFFD—such as regularization techniques~\cite{li2017learning,hou2019learning,kirkpatrick2017overcoming}, parameter-isolation strategies~\cite{rusu2016progressive,xu2018reinforced,verma2021efficient}, and replay methods~\cite{rebuffi2017icarl,aljundi2019online,chaudhry2020using}—these general approaches have not been expressly tailored for face forgery detection. This oversight often culminates in a performance that is less than ideal. To remedy this, we identify two distinct characteristics that differentiate CFFD from conventional continual learning: (1). Face forgery detection is inherently a binary classification task. It is mainly concerned with differentiating an array of fake faces, implying that features of real faces tend to exhibit more uniformity. (2). The differences between real and forged faces are subtle, putting the onus on the detection network to identify the discriminative differences effectively.

In this paper, we propose a novel Historical Distribution Preserving (HDP) framework that preserves the historical distributions of real and forged faces. The main challenge is \textit{how to efficiently preserve historical forgery distribution and the discriminative differences from real faces?} Motivated by the universal adversarial perturbation (UAP), which can be regarded as a \textit{feature} that captures the main direction of image-space that affects the model decision~\cite{jetley2018friends,zhang2020understanding}, we propose a new perspective of continual learning that uses the UAP as the discriminative feature of forged faces relative to real faces. Specifically, instead of storing redundant forged data, we only store a single UAP generated by the historical model. And when dealing with new forgery attacks, we combine the real faces and the UAP as the pseudo-forged faces to simulate historical forgery distribution.
Such faces can not only reserve the discriminative feature of the forgery but also protect privacy from being violated. Feature-based knowledge distillation is further employed to maintain the distribution of real and pseudo-forged faces across different models to reduce the domain gap between each stage. The inherent uniformity of real-face features allows for easier maintenance of the real distribution and further ensures the restoration of historical distribution by the pseudo-forged samples.


Notably, our methodology demands minimal storage overhead and is seamlessly integrable with any classification-centric face forgery detection framework within the proposed CFFD paradigm. This bears significant ramifications for real-world implementations, where efficiency and adaptability are paramount. Extensive experiments on the CFFD benchmark show that our method not only outperforms state-of-the-art competitors but also exhibits exceptional resilience against continually evolving forgery techniques. In summary, our main contributions are as follows:
\begin{itemize}
    \item 
    We first exploit the Universal Adversarial Perturbation (UAP) into continual learning as the discriminative historical feature instead of storing samples.
    \item 
    We propose a novel Historical Distribution Preserving (HDP) framework in the Continual Face Forgery Detection task to preserve the distributions of real and forgery faces and their discriminative distribution difference, thus mitigating catastrophic forgetting. 
    \item We provide a benchmark with three evaluation protocols for the CFFD task. Extensive experiments and visualizations demonstrate the effectiveness of our method. 

\end{itemize}

\section{Related Work}\label{sec2}

\subsection{Face Forgery Detection}

Face forgery detection is mainly to identify whether the input face is forged or not.
Early studies use hand-crafted features to seek the artifacts in forged faces~\cite{matern2019exploiting,yang2019exposing}. Subsequently, deep learning-based works achieve better performance by extracting high-level features for classification. 
For example, some works~\cite{stehouwer2019detection,zhao2021multi} highlight the manipulated regions through attention mechanism, while others~\cite{chen2021local,zhao2021learning} leverage the self-consistency of local regions as forged clues. Although these methods achieve extraordinary performance on seen forgery attacks, they suffer from significant performance degradation when testing under new forgery attacks.
Subsequently, some methods have attempted to learn generalized representation. 
Face X-ray~\cite{li2020face} conducts self-supervised learning driven by simulation of the blending traces. LTW~\cite{sun2021domain} uses meta-learning to reweight samples and provide gradient regularization. DCL~\cite{sun2021dual} controls the intra-class variance to preserve the transferability via contrastive learning.
However, their performance on unseen attacks is still far from practical application.
With the continuous emergence of new forgery methods, it is difficult to obtain all forgery samples with various manipulation techniques at once. And performing finetune directly based on historical training data often requires much time and storage costs.
To address these issues, We focus on a new yet practical face forgery detection task, named Continual Face Forgery Detection (CFFD)\cite{li2023continual}, which enables efficient learning to deal with continuous new attacks.
\begin{figure*}[t!]
    \begin{center}
    \includegraphics[width=1\textwidth]{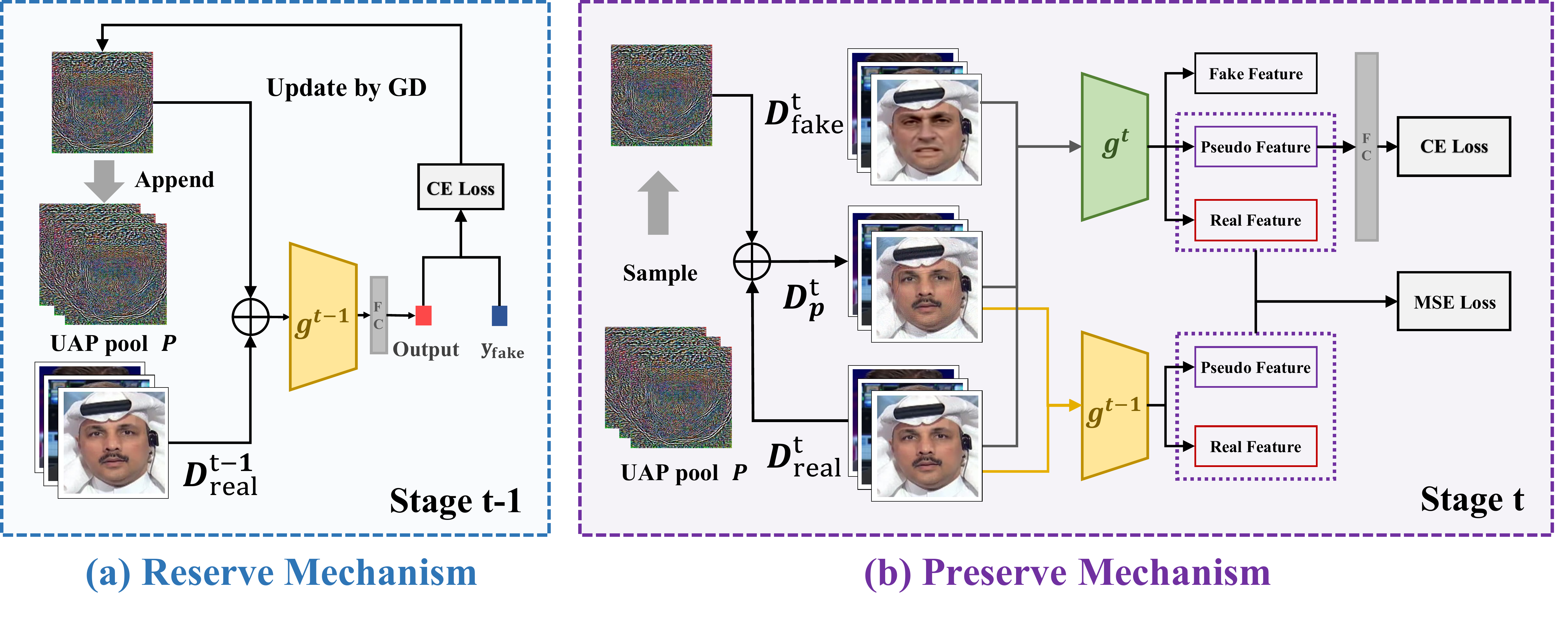}
    \end{center}
       \caption{Overview of the proposed Historical Distribution Preserving framework for CFFD task. Subfigure (a) elucidates the reservation mechanism and subfigure (b) details the preservation mechanism.}
    \label{fig:main}
\end{figure*}
\subsection{Continual Learning}
Current continual learning methods can be taxonomized into three major categories\cite{delange2021continual}.
1) \textit{Parameter isolation} methods intend to assign separate components for each task. These works typically require a task oracle~\cite{delange2021continual} that matches the corresponding dynamic layer to a special task, which increases the network complexity and is not conducive to practical deployment.
2) \textit{Regularization-based} methods introduce an extra regularization term in the loss function to consolidate previous knowledge during training new tasks. 
Specifically, LwF~\cite{li2017learning} uses the output of previous models as soft-label and treats the regularization term as knowledge distillation. 
EWC~\cite{kirkpatrick2017overcoming} and MAS~\cite{aljundi2018memory} try to estimate a distribution over the model parameters.
Other regularization-based methods estimate a distribution over the model parameters such as EWC~\cite{kirkpatrick2017overcoming} and MAS~\cite{aljundi2018memory}. 
Though these approaches alleviate catastrophic forgetting to some extent, they may yield suboptimal performance when faced with challenging settings or complex datasets~\cite{wang2021learning}.
3) \textit{Replay-based} methods preserve the samples of previous tasks in a memory bank and replay them when learning the current task.
The most classic method is iCaRL~\cite{rebuffi2017icarl}, which stores a subset of exemplars based on the distance of the class means center in the feature space.
Recently, some works~\cite{buzzega2020dark,chaudhry2020using,wu2019large} combine the knowledge distillation regularization with a memory bank to further avoid overfitting.
Although replay methods can alleviate catastrophic forgetting well, the performance is vulnerable to the size of buffers, and the extra memory bank may also bring data privacy leakage and storage burden. Differently, 
we propose a new perspective of continual learning that introduces the universal adversarial perturbation (UAP) to simulate historical distribution. Remarkably we store only one UAP generated by the historical model without any additional rehearsal buffers, while still achieving better performance than traditional replay methods.

\subsection{Universal Adversarial Perturbation}
Universal Adversarial Perturbation (UAP)~\cite{moosavi2017universal} is a specially designed noise that can fool the deep learning model with high probability. Different from the per-instance adversarial examples~\cite{szegedy2013intriguing}, UAP needs to compute only a single perturbation vector to fool all the images. Moosavi-Dezfooli~\emph{et al.}~\cite{moosavi2017universal} first formulate and generate UAP via iterative deepfool~\cite{moosavi2016deepfool}. 
Meanwhile, the interpretability of UAP has drawn lots of attention in the computer vision community. Moosavi-Dezfooli~\emph{et al.}~\cite{moosavi2017robustness} theoretically proved that the existence of a low dimension subspace that captures the correlation between the decision boundary of the target model is the reason why UAP works~\cite{chaubey2020universal}. Furthermore, Jetley~\emph{et al.}~\cite{jetley2018friends} reveals the consistency in model performance and robustness. Zhang~\emph{et al.}~\cite{zhang2020understanding} demonstrate that the UAP themselves contains features independent of the images to attack. Inspired by their seminal analysis, our method utilizes UAP and the corresponding data to approximate the previous feature distribution and replay it when training the new task.

\section{Method}\label{sec3}
\subsection{Problem Definition}

The goal of Continual Face Forgery Detection (CFFD) is to develop a unified detector from a sequence of face data encompassing different forgery types and domains. Formally, we denote the training data available at the $t$-th stage as $D^t = \{D_{\text{real}}^t, D_{\text{fake}}^t\}$, where $D_{\text{real}}^t$ and $D_{\text{fake}}^t$ represent the incoming real and forged faces, respectively. The face forgery detection model for the $t$-th stage, $f^t$, is trained exclusively from the dataset $D^t$. Data from the previous stages, $\{ D^t\}_{t=1}^{t-1}$, are no longer accessible. This framework is motivated by two primary considerations: first, it sidesteps the storage overhead and the associated computational challenges posed by a vast archive of historical data; second, privacy constraints might make historical face data unavailable. 
For evaluation, the model $f^t$ is tested against all previously identified forgery attacks, ensuring that its robustness and adaptability are rigorously assessed across a comprehensive range of scenarios.


\subsection{Base Training}
In this section, we outline the fundamental strategy for training the face forgery detection model. Practically, our proposed approach can be incorporated into any State-Of-The-Art (SOTA) classification-based method.
For the sake of clarity, we employ the widely recognized binary classifier model~\cite{dolhansky2020deepfake}, using cross-entropy loss to optimize the model on dataset $D$. This is expressed as:
\begin{equation}
    L_{ce} = -\frac{1}{|D|}\sum_{(x,y)\in D} y\log(f(x)) + (1-y)\log(1-f(x)),
    \label{equation:eq0} 
\end{equation}
where $x \in \mathbb{R^{ C \times H \times W}}$ and $y \in \{0,1\}$ denote the images and their corresponding labels (real/fake) from the dataset $D$, respectively. At the $t$-th stage, if we employ cross-entropy loss directly on $D^t = \{D_{\text{real}}^t, D_{\text{fake}}^t\}$, the model $f^t$ will recognize the forgery types it has been exposed to but will likely forget about earlier forgery techniques. Hence, the primary challenge of CFFD lies in preserving the historical forgery distribution and the discriminative differences between real and forgery faces.


\subsection{Overall Framework}
For continual face forgery detection, we introduce a Historical Distribution Preserving (HDP) framework. This framework tackles the challenge of catastrophic forgetting with two main components: the Reserve Mechanism and the Preserve Mechanism. The operational flow of the HDP is depicted in Fig.\ref{fig:main}.
Within the Reserve Mechanism, we utilize the Universal Adversarial Perturbation (UAP)~\cite{moosavi2017universal} as the discriminative feature distinguishing forged faces from real ones. Specifically, we generate the UAP from the proficiently trained model $f^{t-1}$, encapsulating the most discerning features of forgery attacks up to the $t-1$ stage and storing them in the UAP pool.
Transitioning to a new attack at the $t$ stage, we deploy the Preserve Mechanism. This strategy begins by melding real faces with the generated UAPs, simulating the distribution of forgery attacks from previous stages. Subsequently, this synthesized data is trained in tandem with the fresh dataset. Furthermore, we conduct feature-wise knowledge distillation between pseudo and real features across the current and the last step model to maintain the stability of the distribution in all stages.

\subsection{Reserve Mechanism}

The primary objective of this mechanism is to preserve the historical forgery distribution by a combination of Universal Adversarial Perturbation (UAP) and real data. As highlighted in prior studies~\cite{zhang2020understanding,jetley2018friends}, UAP embodies a feature that consistently imparts directional perturbations across class boundaries, regardless of the diverse individual appearances. Furthermore, Jetley~\emph{et al.}~\cite{jetley2018friends} elucidated the profound correlation between specific directions and class identity—essentially, the network discerns these directions as features inherently associated with class identities. Consequently, we employ UAP as a distinctive feature to differentiate forged faces from real ones, combining it with real faces to synthesize pseudo-forged samples.
For a model \(f^{t-1}\) trained during the \(t-1\) phase, which incorporates a feature extractor \(g^{t-1}\), an FC layer, and real subset samples \(x_{\text{real}}^{t-1} \in D_{\text{real}}^{t-1}\) labeled \(y_{\text{real}}=0\), our goal is to ascertain perturbation vectors \(p^{t-1} \in \mathbb{R^{C \times H \times W}}\) capable of misleading the model into misclassifying real images as fake. Analogous to \cite{moosavi2017universal}, we target a vector \(p^{t-1}\) that conforms to:
\begin{equation}
    \begin{split}
    \text{Pred}(f^{t-1}&(x_{\text{real}}^{t-1}+p^{t-1})) \neq y_{\text{real}}\\
    &\textbf{s.t.} ||p^{t-1}||_{\infty} \leq \epsilon,
    \end{split}
\end{equation}
where \(\text{Pred}\) translates the logit to a prediction value (i.e., \(0\) for real and \(1\) for fake), and \(\epsilon\) modulates the magnitude of the perturbation vector.
We denote the pseudo-forged samples \(x^{t-1}_{p}\) as:
\begin{equation}
    x^{t-1}_{p} = x_{\text{real}}^{t-1} + p^{t-1}.
\end{equation}
Contrary to the native UAP approach~\cite{moosavi2017universal} that deploys deepfool~\cite{moosavi2016deepfool} to ascertain perturbations defining the decision boundary, our strategy interprets UAP as a potent feature. Taking a cue from~\cite{zhang2020understanding}, we utilize the gradient descent technique to refine \(p^{t-1}\) by minimizing the entropy between pseudo-forged samples \(x^{t-1}_{p}\) and the fake label \(y_{\text{fake}}=1\). This process is encapsulated as:
\begin{equation}
    p^{t-1} = p^{t-1} - \alpha*\text{sgn}(\nabla_{p^{t-1}}\log(f^{t-1}(x^{t-1}_{p}))),
    \label{equation:eq1}
\end{equation}
where \(\text{sgn}\) represents the symbolic function, and \(\alpha\) signifies the learning rate of \(p^{t-1}\). The perturbation is reiterated until \(x^{t-1}_{p}\) transgresses the decision boundary. The algorithm concludes when the number of successfully modified prediction outcomes surpasses the predetermined threshold \(\sigma\).
Upon deriving the UAP \(p^{t-1}\), we integrate it into the UAP pool \(P\) in anticipation of the rehearsal phase, which substantially minimizes computational strain and storage encumbrance.

\subsection{Preserve Mechanism}

Upon reversing the historical distribution, the subsequent step is to retain the forgery distribution by integrating them into the ongoing training process. Specifically, during the $t$-th stage of training, entropy optimization is employed for the pseudo-forged samples paired with their fake labels. This introduces a regularization term, ensuring the distribution of the pseudo-forged samples is duly preserved.

During the training of model $f^t$ with the dataset $D^t$, UAP from stages up to $t-1$, represented by $\{p^n\}_{n=1}^{t-1}$, is sequentially sampled from the UAP pool $P$ before each training iteration. Following this, the pseudo-forged dataset $D^n_{p}$ is produced via pixel-wise summation of $p^n$ and $D_{real}^t$, expressed as $D^n_{p} = p^n + D_{real}^t$, associated with the label $y_{fake}=1$. The entropy of $D^n_{p}$ and the fake label $y_{fake}$ is then computed as:
\begin{equation}
    E^n = - \frac{1}{|D^n_{p}|}\sum_{x^n_{p}\in D^n_{p}}\log(f^t(x^n_{p})).
    \label{equation:eq5}
\end{equation}

\begin{algorithm}[t]
    \label{algorithm}
    \caption{Historical Distribution Preserving}\label{algorithm}
    \begin{algorithmic}[1]
        \Require{Dataset $D = \{D_{real}^t, D_{fake}^t\}_{t=1}^T$, initial model $f^0$, UAP pool $P$.}
        
        \Ensure{ $P\leftarrow$\{\}}
        
            \For{\{$D^t_{real}$,$D^t_{fake}$\} $\in$ $D$}
            \If{$t==1$} 
                \State Optimize $f^0$ with Eq.\ref{equation:eq0} to obtain $f^{1}$.
            \Else
                \State Optimize $f^{t-1}$ with Eq.\ref{equation:final} to obtain $f^{t}$.
            \EndIf
            \State Generate UAP $p^{t}$ for model $f^{t}$.
            \State Append \(p^{t}\) into the \(P\).
            \EndFor
                
        \State\textbf{Return model $f^{T}$}
        
    \end{algorithmic}
    \label{alg:main2}
\end{algorithm}

To quantify the overall training entropy \(E^t\) at the t-th stage, one must consider the cumulative effect of various entropies \(E^n\) that have been sequentially sampled from the UAP pool. Thus, the overall training entropy is not just influenced by the current stage but is an accumulation of the influences from prior stages as well. 
This ensures that the distribution of forgery attacks from the previous stage is perpetuated. Nevertheless, given that parameters undergo dynamic optimization, the consistency of the pseudo-forged distribution might be compromised. Addressing this concern, we implement feature-wise knowledge distillation, utilizing the prior feature extractor $g^{t-1}$. This maintains distillation based on the pseudo-forged samples $x^{t}_{p}$, and is formalized as:
\begin{equation}
    L^t_{p} =  \frac{1}{|D^n_{p}|}\sum_{x_{p}\in D^n_{p}}||g^t(x_{p}) - g^{t-1}(x_{p})||^2.
    \label{equation:eq6}
\end{equation}

It's noteworthy that in face forgery detection, the emphasis is predominantly on discerning forgery traces rather than differentiating between real and fake facial content. In practical applications, genuine images exhibit a relative homogeneity compared to their forged counterparts. Nonetheless, new forgery techniques can alter the real distribution, influencing the simulation of pseudo-forged samples. To counteract this, we emphasize maintaining the continuity of real distillation.
Following the same paradigm as $L_p$, a single-set feature-based knowledge distillation is applied. Without any compromise to generality, let $g^t$ be the feature extractor for the model $f^t$. Here, the previous feature extractor, $g^{t-1}$, is regarded as the teacher, utilizing the mean-squared function as the regularization term. This can be articulated as:
\begin{equation}
    L^t_{r} = \sum_{x\in D_{real}^t}||g^t(x) - g^{t-1}(x)||^2.
    \label{equation:eq7}
\end{equation}
By regularizing each training stage using the model from the preceding phase, the real sample distribution remains consistent, ensuring stability.


\subsection{Loss Function}
In summary, when new forgery attacks arise at the $t$ stage, the overall loss function for our proposed method of preserving historical distribution is defined as the sum of the base training loss and the aforementioned distribution-preserving optimizations. This can be expressed as:
\begin{equation}
L^t = L^t_{ce} + E^t + \beta(L^t_{r}+L^t{_p}),
\label{equation:final}
\end{equation}
where $\beta$ is a hyper-parameter that weighs the regularization terms. To offer a clearer understanding of our method, we provide the pseudo-code of the HDP in Alg~\ref{alg:main2}.

\section{Experiment}\label{sec4}
\subsection{Experimental Setting}
\noindent\textbf{Datasets.}
We conduct experiments based on four challenging datasets:
FaceForensics++~\cite{rossler2019faceforensics++} (FFpp) is a widely-used dataset containing four different face synthesis methods, including two deep learning-based methods Deepfakes (DF) and NeuralTextures (NT) and two graphics-based approaches Face2Face (FF) and FaceSwap (FS).
Celeb-DF~\cite{li2020celeb} is another Deepfake dataset that collects real source videos from Youtube and generates forgery videos using an improved DeepFake synthesis method, resulting in a higher quality of forgery samples.
DFDC~\cite{dolhansky2020deepfake} is a large-scale dataset with various manipulated methods and backgrounds.
Wild-Deepfake~\cite{zi2020wilddeepfake} is a face forgery dataset where all videos are obtained from the internet, thus it has various synthesis methods, identities, and image qualities.

\noindent\textbf{CFFD Benchmark.}
We introduce a novel CFFD benchmark that encompasses three protocols, each escalating in difficulty.

\textbf{Protocol 1:} This protocol comprises four fake subsets from FFpp. Though these subsets maintain consistency in context and individual identity, they vary in their manipulation types. The core objective of Protocol 1 is to emulate the continuous emergence of novel attack strategies.

\textbf{Protocol 2:} Diversifying the range, this protocol draws from four fake datasets, namely FFpp, Celeb-DF, Wild-Deepfake, and DFDC. When juxtaposed with Protocol 1, Protocol 2 exhibits a wider array of attack types and more pronounced domain gaps. The real samples in this protocol differ at each stage, which elevates its difficulty. This intricacy, combined with its expansive applicability, positions Protocol 2 as particularly challenging.

\textbf{Protocol 3:} Tailored to assess the ability to handle extended sequences of evolving real and fake imagery, this protocol unfolds across ten stages. It utilizes subsets from FFpp and evenly apportions the datasets from Celeb-DF, Wild-Deepfake, and DFDC. This layout accentuates the risk of catastrophic forgetting. Between the three protocols, Protocol 3 most accurately mirrors the complexities of real-world scenarios, typified by extended sequences and multi-domain data.
For an in-depth exploration of the three protocols, please refer to the supplementary materials.

\begin{table*}[!t]
    \renewcommand\arraystretch{1.1}
    \centering
    
    \vspace{1pt}
    \resizebox{\textwidth}{!}{
    \begin{tabular}{c|c|cc|cc|cc|cc|cc|cc}
        \hline
        \multirow{2}*{Method} &Buffer &\multicolumn{2}{c|}{Deepfakes}&\multicolumn{2}{c|}{Face2Face} & \multicolumn{2}{c|}{FaceSwap} & \multicolumn{2}{c|}{NeuralTextures} &\multicolumn{2}{c|}{AVG}&\multicolumn{2}{c}{PRE}\\
        \cmidrule{3-14}
        &/Step & ACC& AUC & ACC& AUC& ACC& AUC& ACC& AUC& ACC& AUC& ACC& AUC\\
        \hline
        SFT&0&76.88&91.88&68.15&85.57&59.74&72.78&\underline{95.53}&\textbf{98.43}&75.07&87.16&67.29&88.36\\
        LwF&0&81.51&95.61&72.26&91.81&61.27&81.38&95.48&\underline{98.28}&77.63&91.77&74.09&94.05\\
        MAS&0&86.97&95.14&78.01&86.81&65.89&72.89&90.86&96.15&80.43&87.74&\underline{83.76}&90.75\\
        NCM&0&76.94&92.90&67.67&92.99&61.77&70.75&\textbf{95.66}&96.81&75.53&85.28&69.13&87.73\\

        iCaRL&500&84.46&95.67&81.71&93.28&73.41&87.01&94.70&98.15&83.57&93.52&82.82&93.84\\
        SCR&500&82.85&\underline{96.75}&81.42&\underline{94.79}&62.96&89.49&95.53&97.33&80.69&86.75&75.29&\underline{95.27}\\
        FReTAL&500&\underline{86.95}&94.26&\underline{88.55}&93.65&\underline{88.50}&\underline{94.85}&94.43&98.25&\underline{89.53}&\underline{95.25}&80.07&95.24\\
        DER&500&78.41&87.82&73.32&82.63&71.80&74.24&88.91&93.14&77.86&84.45&80.50&89.10\\
        DER++&500&81.20&94.10&74.93&88.90&70.15&76.20&90.23&97.40&79.12&88.64&80.60&89.09\\
        Co2L&500&84.91&94.81&76.50&86.81&64.99&73.95&94.36&97.64&80.19&88.30&77.52&89.00\\
        \hline
        HDP&0&\textbf{93.10}&\textbf{98.17}&\textbf{91.90}&\textbf{96.86}&\textbf{93.47}&\textbf{97.52}&94.81&97.12&\textbf{93.32}&\textbf{97.41}&\textbf{94.94}&\textbf{98.50}\\
        \hline
        Joint&-&96.99&99.75&96.35&99.12&96.63&99.30&94.04&97.84&96.00&99.00&-&-\\
        \hline
    \end{tabular}
    
    }
    \caption{Quantitative results on protocol1 in terms of ACC and AUC. We evaluate the last model on all the previous test sets. The buffer represents the memory bank budget save the previous images per stage.}
    \label{table:protocol1}
    
\end{table*}

\noindent\textbf{Evaluation Metrics.}
Our primary evaluation metrics encompass the accuracy score (ACC) and the area under the receiver operating characteristic curve (AUC). Both ACC and AUC are standard measures to evaluate the performance of classification models. While ACC measures the proportion of correctly classified instances out of the total instances, AUC provides an aggregate measure of the model's performance across all possible classification thresholds.

In the context of the CFFD benchmark, we employ two primary metrics:

\textbf{AVG (Average Performance)} represents the performance of the \textit{last model} over each individual task. By evaluating the last model across the various tasks, we aim to understand how well the continually updated model performs on different challenges, especially the most recent ones. In mathematical terms, AVG calculates the mean of the performance metric (be it ACC or AUC) of the last model on each separate task.

\textbf{PRE (Performance on Previous Tasks)} is designed to gauge the model's resilience against catastrophic forgetting. Catastrophic forgetting refers to the model's tendency to overwrite previous knowledge when learning new information. PRE, therefore, calculates the average performance metric of the model on all previous tasks, relative to the current task. A high PRE score indicates that the model retains its proficiency on prior tasks even after being updated with new tasks, showcasing its resistance to catastrophic forgetting.


\begin{figure*}[h]
    \begin{center}
    \includegraphics[width=1.0\textwidth]{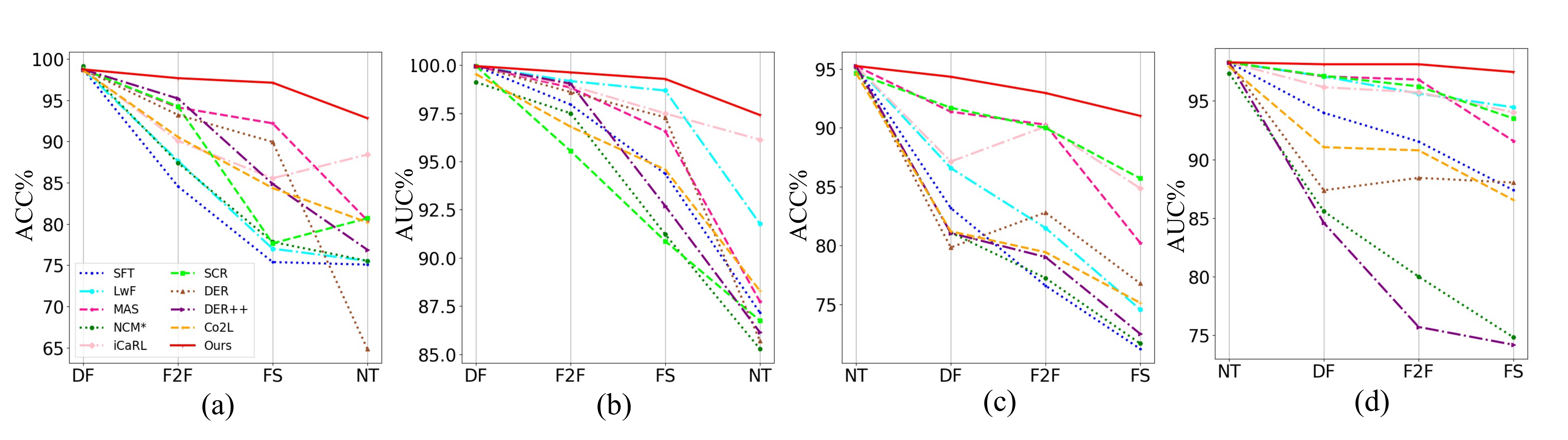}
    \end{center}
       \caption{Illustration of the non-forgetting evaluation of each stage on protocol 1. (a) and (b) show the trend of average AUC and ACC for the previous dataset during the training process for order 1, respectively. (c) and (d) represent the order2.
       }
    \label{fig:plot}
\end{figure*}


\noindent\textbf{Implementation Details.}
To ensure a fair comparison, all comparative methods are aligned to the following settings, unless specified otherwise. We utilize the EfficientNet-b4~\cite{tan2019efficientnet}, pre-trained on ImageNet~\cite{deng2009imagenet}, as the backbone for our face forgery detection model. DSFD~\cite{li2019dsfd} serves as our face detector across all datasets. All input face images are resized to a resolution of $224\times224$.
For model training, we employ the Adam optimizer. Key parameters are set as follows: weight decay is $1e-5$, the learning rate is $0.001$, and the batch size is fixed at 64. While generating the UAP, we adjust the norm of perturbation ($\epsilon$) to $0.15$, set $\alpha$ at $0.0001$, and determine the successful threshold ($\sigma$) to be $0.8$. The complete framework is realized using PyTorch and is run on a single NVIDIA A-100 GPU.

\subsection{Quantitative Results}
\noindent\textbf{Comparing Methods.}
We benchmark our method against several state-of-the-art continual learning techniques: including 1) Sequence fine-tune (SFT): Directly fine-tune model on new task dataset without extra operation; 2) Learning without forgetting (LwF)~\cite{li2017learning}: A regularization-based technique that retains previous knowledge via knowledge distillation. 3) Memory Aware Synapses (MAS)~\cite{aljundi2018memory}: Another regularization-based strategy that forms the regularization term using the previous model parameters and their corresponding importance weight. 4) Nearest Class Mean Classifier for real sample ((NCM*)~\cite{zhang2020understanding}: This method adjusts the unified classifier to our setting, constraining the real cluster centers and using the NCM classifier in place of the softmax-fc layer.
5) iCaRL~\cite{rebuffi2017icarl}: A well-known replay method that chooses to best approximate class means in the feature space and stores them in a memory bank.
6) Supervised Contrastive Replay (SCR)~\cite{mai2021supervised}: A recent replay technique that merges supervised contrastive learning with the NCM classifier.
7) Continual Representation using Distillation (FReTAL)~\cite{kim2021fretal}: The first approach that considers continual learning settings within the face forgery detection task.
8) Dynamically Expandable Representation (DER) and DER++~\cite{yan2021dynamically}: A unique two-stage learning approach that employs dynamically expandable representation for more effective incremental concept modeling.
9) Contrastive Continual Learning (Co2L)~\cite{cha2021co2l}: This method learns representations through the contrastive learning objective and preserves these representations using a self-supervised distillation step.
For all the replay methods mentioned above, we set the memory buffer size to hold 500 forgery images for each task.

\noindent\textbf{Quantitative Results on Protocol1.}
Since sequence order is agnostic in continual learning, two orders are evaluated in protocol 1. \textit{Order1} is represented by \textbf{DF$\rightarrow$F2F$\rightarrow$FS$\rightarrow$NT} and \textit{Order2} is \textbf{NT$\rightarrow$DF$\rightarrow$F2F$\rightarrow$FS}. The adjacent attack types in Order1 are more alike, in contrast to Order2, where the manipulation methods between stages are distinct.
The results of \textit{Order1} are shown in Tab.~\ref{table:protocol1}, while that of \textit{Order2} are presented in the Appendix due to paper space limitations. In
In Tab.\ref{table:protocol1}, it's evident that our proposed approach significantly surpasses the competing methods in terms of both ACC and AUC metrics.
Relative to the SFT, which grapples with pronounced catastrophic forgetting, our method sees approximately a $27\%$ enhancement in the PRE metric. This uplift primarily stems from the introduced historical distribution preservation strategy.
Further, the performance of regularization-centric strategies, like LwF and MAS, is hampered by the pronounced domain gap between Celeb-DF and NeuralTextures — a challenge our method can adeptly navigate. Unlike existing replay strategies that demand large buffer sizes, our method requires storing only one UAP per task, giving us an advantage in preventing forgetting with minimal storage overhead.

To further show the ability to combat forgetting, we illustrate the average performance of the previous task and current task during each training process on \textit{order1} and \textit{order2}, respectively. As shown in Fig.~\ref{fig:plot} (a) and (b), our methods can achieve state-of-the-art results under different stages on both ACC and AUC. Fig.~\ref{fig:plot} (c), (d) summarizes the results on the \textit{order2}. Similar to the \textit{order1}, our method achieves the SOTA performance of AVG and PER compared with other methods on \textit{order2}.
Specifically, when facing the more difficult order, the ACC of AVG and PRE still obtain 20\% and 25\% improvement compared with SFT. The detailed results of  Order2 are presented in the Appendix. 
The above results demonstrate the effectiveness and robustness of our approach to different task sequencing.

\begin{table*}[!t]
    \renewcommand\arraystretch{1.1}
    \centering
    
    \vspace{1pt}
    \resizebox{\textwidth}{!}{
    \begin{tabular}{c|c|cc|cc|cc|cc|cc|cc}
        \hline
        \multirow{2}*{Method} &Buffer &\multicolumn{2}{c|}{FFpp}&\multicolumn{2}{c|}{Celeb-DF} & \multicolumn{2}{c|}{Wild-Deepfake} & \multicolumn{2}{c|}{DFDC} &\multicolumn{2}{c|}{AVG}&\multicolumn{2}{c}{PRE}\\
        \cmidrule{3-14}
        &/Step & ACC& AUC & ACC& AUC& ACC& AUC& ACC& AUC& ACC& AUC& ACC& AUC\\
        \hline
        SFT&0&85.02&93.88&81.75&90.17&72.59&83.27&\underline{87.74}&\underline{95.67}&81.77&90.82&79.92&89.38\\
        LwF&0&87.08&96.64&83.82&90.69&73.77&84.85&87.12&94.84&82.94&91.75&82.93&92.84\\
        MAS&0&85.45&97.16&81.17&89.29&76.91&81.03&87.03&90.10&81.86&91.64&82.22&92.00\\
        NCM&0&85.83&93.44&83.76&90.22&77.01&83.72&86.56&93.76&83.29&90.28&80.38&90.14\\
        
        iCaRL&500&85.50&95.62&85.07&92.77&74.87&84.49&87.81&95.46&83.38&92.08&84.21&93.15\\
        SCR&500&86.90&96.46&\underline{86.01}&92.99&76.71&85.59&88.19&95.63&84.45&92.66&83.30&93.97\\
        FReTAL&500&\underline{90.67}&\underline{96.78}&85.62&\underline{94.26}&76.58&86.34&87.12&95.52&84.99&93.22&83.70&93.43\\
        DER++&500&87.89&94.40&84.92&91.05&73.58&85.48&87.29&95.35&83.42&91.57&83.35&93.77\\
        Co2L&500&87.69&95.77&84.57&90.08&73.79&81.20&86.30&91.77&83.08&89.70&83.07&92.57\\
        \hline
        HDP&0&87.81&96.48&86.91&93.20&\underline{78.03}&\underline{87.21}&87.51&95.05&\underline{85.63}&\underline{93.23}&\underline{86.99}&\underline{94.09}\\
        HDP&\textbf{50}&\textbf{91.57}&\textbf{98.26}&\textbf{87.68}&\textbf{94.98}&\textbf{80.31}&\textbf{87.91}&\textbf{88.86}&\textbf{95.77}&\textbf{86.53}&\textbf{94.03}&\textbf{88.57}&\textbf{95.31}\\
        \hline
        Joint&-&96.05&99.67&98.53&99.89&83.39&90.40&87.56&96.60&91.88&96.64&-&-\\
        \hline
    \end{tabular}
    }
    \caption{Quantitative results on protocol2. We evaluate the last model on the test sets of all tasks in terms of ACC and AUC. The training order is FFpp $\rightarrow$ Celeb-DF$\rightarrow$ WildDeepfake$\rightarrow$ DFDC. The buffer represents the memory bank budget save the previous images per stage. Underline indicate the sub-optimal results.}
    \label{table:protocol2}
    
\end{table*}

\noindent\textbf{Quantitative Results on Protocol2.}
In Protocol 2, which is characterized by a substantial domain gap, the input order is set as \textbf{FFpp$\rightarrow$Celeb-DF$\rightarrow$WildDeepfake$\rightarrow$DFDC}. As seen in Tab.~\ref{table:protocol2}, our method notably surpasses other comparative methods in terms of both ACC and AUC for AVG and PRE metrics. Specifically, the ACC for PRE registers at 86.99\%, marking a 7\% enhancement from the baseline SFT. Our technique achieves state-of-the-art (SOTA) performance without any buffer, distinguishing it from other methods reliant on buffers.
In contrast to buffer-based strategies like FReTAL, which reserves 500 images per stage, our method attains a 3\% boost in the PRE metric without retaining any historical data. Additionally, when we augment our HDP to accommodate historical data with a buffer size of 50 images per stage, our method consistently outperforms competitors, even with ten times less storage. This adaptability underscores our method's efficacy.
These results attest to our method's prowess in accurately approximating the forgery distribution and retaining knowledge adeptly, even in environments with pronounced domain gaps. Importantly, our method doesn't just alleviate catastrophic forgetting but also matches the performance levels of contemporary forgery attacks (from the latest stage) seen in other methodologies.




\begin{table}[!t]

    \centering
    \renewcommand\arraystretch{1.1}
    \resizebox{1.0\columnwidth}{!}{
    \begin{tabular}{c|c|c|c|c|c}
        \hline
        \multirow{2}*{Method}&Buffer/&\multicolumn{2}{c|}{AVG}&\multicolumn{2}{c}{PRE}\\
        \cmidrule{3-6}
        &Stage&ACC&AUC&ACC&AUC\\

        \hline
        SFT&0&72.51&80.30&68.90&76.64\\
        LwF&0&74.32&83.81&70.99&78.58\\
        MAS&0&73.88&83.56&69.89&78.92\\
        NCM&0&72.98&81.25&70.37&77.38\\
        iCaRL&500&74.17&83.50&72.77&79.94\\
        SCR&500&75.66&85.61&72.79&80.69\\
        FReTAL&500&76.18&84.97&73.00&81.53\\
        DER++&500&73.87&82.21&71.19&78.98\\
        Co2L&500&75.66&84.43&72.95&81.43\\
        \hline
        HDP&0&\underline{77.17}&\underline{85.94}&\underline{74.10}&\underline{82.01}\\
        HDP&\textbf{50}&\textbf{80.15}&\textbf{87.97}&\textbf{75.47}&\textbf{83.82}\\
        \hline
    \end{tabular}
    }
    \vspace{4pt}
    \caption{Quantitative results on protocol3. We evaluate the last model on the test sets of all tasks in terms of ACC and AUC. The buffer represents the memory bank budget save the previous images per stage. Underline indicate the sub-optimal results.}
    \label{table:protocol3}
\end{table}
\begin{table*}[!h]
    \renewcommand\arraystretch{1.1}
    \centering
    
    \vspace{1pt}
    \resizebox{\textwidth}{!}{
    \begin{tabular}{c|cc|cc|cc|cc|cc|cc}
        \hline
        \multirow{2}*{Method}  &\multicolumn{2}{c|}{Deepfakes}&\multicolumn{2}{c|}{Face2Face} & \multicolumn{2}{c|}{FaceSwap} & \multicolumn{2}{c|}{NeuralTextures} &\multicolumn{2}{c|}{AVG}&\multicolumn{2}{c}{PRE}\\
        \cmidrule{2-13}
         & ACC& AUC & ACC& AUC& ACC& AUC& ACC& AUC& ACC& AUC& ACC& AUC\\
        \hline
        EN-B4&76.88&91.88&68.57&85.57&59.74&72.78&\textbf{95.53}&\textbf{98.43}&75.07&87.16&67.29&88.36\\
        EN-B4+HDP&\textbf{93.10}&\textbf{98.17}&\textbf{91.90}&\textbf{96.86}&\textbf{93.47}&\textbf{97.52}&94.81&97.12&\textbf{93.32}&\textbf{97.41}&\textbf{94.94}&\textbf{98.50}\\
        \hline
        F3-Net&80.94&93.04&69.71&82.70&63.55&74.03&\textbf{94.55}&97.89&77.18&86.91&69.33&89.57\\
        F3-Net+HDP&\textbf{91.65}&\textbf{96.26}&\textbf{89.06}&\textbf{94.70}&\textbf{90.99}&\textbf{96.15}&94.21&\textbf{97.97}&\textbf{91.49}&\textbf{96.22}&\textbf{93.51}&\textbf{97.77}\\
        \hline
        MAT&77.13&92.30&67.75&83.33&66.75&73.49&95.20&\textbf{98.54}&76.81&86.90&71.96&90.72\\
        MAT+HDP&\textbf{92.55}&\textbf{97.14}&\textbf{90.76}&\textbf{95.26}&\textbf{94.23}&\textbf{97.94}&\textbf{95.23}&98.07&\textbf{93.13}&\textbf{96.84}&\textbf{94.36}&\textbf{98.65}\\
        \hline
        SIA&78.52&93.68&71.29&87.21&61.15&75.03&95.44&\textbf{98.40}&76.60&88.58&72.39&91.28\\
        SIA+HDP&\textbf{92.33}&\textbf{98.01}&\textbf{91.11}&\textbf{95.59}&\textbf{93.01}&\textbf{97.31}&\textbf{95.47}&97.39&\textbf{92.98}&\textbf{97.07}&\textbf{95.30}&\textbf{97.58}\\
        \hline
        UIA-VIT&81.43&93.57&70.77&83.51&61.05&71.23&\textbf{95.98}&\textbf{98.25}&77.30&86.64&70.33&90.55\\
        UIA-VIT+HDP&\textbf{93.22}&\textbf{97.52}&\textbf{89.78}&\textbf{95.21}&\textbf{95.07}&\textbf{98.01}&95.22&97.99&\textbf{93.32}&\textbf{97.18}&\textbf{94.81}&\textbf{98.99}\\
        
        \hline
    \end{tabular}
    }
    \caption{Quantitative results on protocol 1 on three SOTA face forgery detection models in terms of ACC and AUC. We evaluate the last model on all the previous test sets. }
    \label{table:sota}
    
\end{table*}

\begin{table}[!t]
    \centering
    \renewcommand\arraystretch{1.1}
    \resizebox{1.0\columnwidth}{!}{
    \begin{tabular}{c|c|c|c|c|c|c}
        \hline
        
        \multirow{2}*{Eq.\ref{equation:eq5}}& \multirow{2}*{Eq.\ref{equation:eq6}} & \multirow{2}*{Eq.\ref{equation:eq7}}&\multicolumn{2}{c|}{AVG}&\multicolumn{2}{c}{PRE}\\
        \cmidrule{4-7}

        &&& ACC &AUC& ACC &AUC\\
        \hline
        &&&75.07&87.16&67.29&88.36\\
        \checkmark&&&77.17&90.33&74.33&92.21\\
        \checkmark&\checkmark&&79.35&92.21&77.39&94.40\\
        &&\checkmark&81.58&93.76&81.19&95.34\\
        \checkmark&&\checkmark&91.79&96.63&90.71&97.11\\
        \hline
        \checkmark&\checkmark&\checkmark&\textbf{93.32}&\textbf{97.41}&\textbf{94.94}&\textbf{98.50}\\

        \hline
    \end{tabular}
    }
    \caption{Ablation study on the influence of different components on protocol1 in terms of ACC and AUC.}
    \label{table:ablation}
\end{table}

\noindent\textbf{Quantitative Results on Protocol 3.}
To rigorously assess our method's performance across longer sequences, we executed comparative tests using the more complex Protocol 3. The sequence followed is \textbf{DF$\rightarrow$Celeb-DF1$\allowbreak\rightarrow$NT$\allowbreak\rightarrow$WildDeepfake1$\allowbreak\rightarrow$FS$\allowbreak\rightarrow$DFDC1$\allowbreak\rightarrow$F2F$\allowbreak\rightarrow$Celeb-DF2$\allowbreak\rightarrow$WildDeepfake2$\allowbreak\rightarrow$DFDC2}
 with '1' and '2' indicating separate portions of the original dataset.
Results for both AVG and PRE, expressed in terms of ACC and AUC for Protocol 3, are detailed in Tab~\ref{table:protocol3}. Evidently, our method attains state-of-the-art (SOTA) results irrespective of buffer size usage.
When pitted against non-buffer reliant strategies like LwF, HDP's PRE outclasses by a 4\% margin in ACC. We also tailored our HDP with a buffer size of 50 samples per stage, akin to our adaptation in Protocol 2. When juxtaposed with methods that utilize buffers of 500 samples, our strategy yields superior outcomes using just 50 samples at each stage. For instance, when gauging in terms of ACC, HDP surpasses FReTAL by 4\% in AVG and 2\% in PRE.
These findings underline the potency of our proposed HDP in extended sequence environments, affirming its prospective utility in real-world applications.


\noindent\textbf{Comparison with Existing Detection Methods.}
To underscore the versatility of HDP, we integrated it with three state-of-the-art (SOTA) face forgery detection methods: F3Net~\cite{qian2020thinking}, Mult-Attentional (MAT)~\cite{zhao2021multi}, SIA~\cite{sun2022information}, and UIA-VIT~\cite{zhuang2022uia}. The results, presented in Tab.~\ref{table:sota}, show that while standalone SOTA methods deliver sub-optimal performance, their efficiency improves notably in terms of both AVG and PRE when combined with the HDP framework. Specifically, integrating HDP into MAT leads to a 23\% improvement in the PRE metric (in terms of ACC), while F3-Net experiences an approximate 16\% enhancement in AVG. These findings not only highlight the robustness of our method but also attest to its adaptability.

\begin{table}[!t]
   \centering
   \renewcommand\arraystretch{1.1}
   \resizebox{\columnwidth}{!}{
   \begin{tabular}{c|c|c|c|c}
       \hline
       
        \multirow{2}*{UAP Attack Rate} &\multicolumn{2}{c|}{AVG}&\multicolumn{2}{c}{PRE}\\
       \cmidrule{2-5}

       & ACC &AUC& ACC &AUC\\
       \hline
       0.6&90.35&96.24&90.33&96.07\\
       0.7&91.03&96.58&92.34&97.30\\
        0.8&\textbf{93.32}&\textbf{97.41}&\textbf{94.94}&\textbf{98.50}\\
       0.9&92.89&97.13&94.10&97.73\\
       1.0&92.37&97.29&93.67&97.25\\
       \hline
   \end{tabular}
   }
   \caption{Ablation study on UAP attack rate.}

   \label{table:ablation}
\end{table}

\subsection{Ablation Study}


\noindent\textbf{Ablation study on components.}
The optimization process of our HDP considers three main equations: Eq.\ref{equation:eq5} which targets the training of pseudo-forged samples, Eq.\ref{equation:eq6}focusing on the distillation of pseudo-forged features, and Eq.\ref{equation:eq7} for real feature distillation. An ablation study was conducted to understand the contributions of each of these equations to the model's performance. The AVG and PRE metrics under Protocol1 in terms of ACC and AUC are shown in Tab.~\ref{table:ablation}. Our observations are multifold: (1). When the regularization of the real distribution is absent, the model witnesses a marked performance decline. This can be attributed to the distribution of pseudo-forged samples, which tends to sway in tandem with shifts in the genuine distribution. This is evident upon contrasting the second, third, and fifth rows. (2). 
An analysis of the fourth row suggests that a constrained focus on real sample distillation yields sub-optimal results. Consequently, the model's proficiency to accurately discern forgery distributions is compromised. (3). Notably, the model's pinnacle of efficacy is attained when all three equations are concurrently and cohesively activated. This substantiates the inherent interdependence of these equations, underscoring the superior results borne from their synergy.

\noindent\textbf{Ablation study on attack rate.}
The generated UAP can represent the discriminative feature of the training set, but there exist many out-of-distribution samples. Thus, the UAP Attack rate, which represents the proportion of successful attacks when generating UAP, is important in our method. To investigate the best attack rate,
we conduct an ablation study on it and reported in Tab.~\ref{table:ablation}.
We varied the attack success rate within [0.6, 0.7, 0.8, 0.9, 1.0], the best results were obtained when the attack rate is 0.8. Since there are some outliers in the training dataset, the attack rate cannot be set too high, otherwise it will cause overfitting. If the attack rate is lower than 0.8, the ability of UAP to represent the distribution will decrease, which will affect the simulation of the distribution and cause knowledge forgetting.




\begin{table}[!t]
    \centering
    \renewcommand\arraystretch{1.1}
    \resizebox{1.0\columnwidth}{!}{
    \begin{tabular}{c|c|c|c|c|c}
        \hline
        Method&Buffer&Epoch&Total Time&AVG&PRE\\
        \hline
        SFT&0&20&2500s&75.07&67.29\\
        LwF&0&20&2900s&77.63&74.09\\
        MAS&0&20&3080s&80.43&83.76\\
        NCM&0&20&3140s&75.53&69.13\\
        iCaRL&500&20&3517s&83.57&82.82\\
        SCR&500&20&3497s&80.69&75.29\\
        FReTAL&500&20&4437s&89.53&80.07\\
        DER++&500&20&4357s&79.12&80.80\\
        Co2L&500&20&4557s&80.19&77.52\\
        \hline
        HDP&0&20&4096s&93.32&94.94\\
        HDP&0&10&\textbf{2136}s&\textbf{91.32}&\textbf{90.63}\\
        \hline
    \end{tabular}
    }
    \vspace{4pt}
    \caption{Analysis of time consumption of each method. Total time represents time spent at each stage. The buffer represents the memory bank budget saved for the previous images per stage. We use ACC as the metric of AVG and PRE.}
    \label{table:time}
\end{table}

\subsection{ Experimental Analysis}

\noindent\textbf{Analysis of Time Consumption.}
To intuitively show the time and memory requirements of each comparison method, we detail the training epoch, memory bank size, and total time per stage alongside AVG and PRE metrics in terms of ACC. From the results in Tab.~\ref{table:time}, it's evident that our method delivers a commendable performance with a significantly reduced memory footprint and training time compared to recent methods like FReTAL, DER++, and Co2L, all of which require 20 training epochs. Moreover, when we reduced our training epochs from 20 to 10, our proposed HDP method still managed to attain state-of-the-art performance, and in terms of time efficiency, it even surpassed SFT. These findings robustly underscore the efficiency and potency of our approach.

\begin{figure*}[t!]
    \begin{center}
    \includegraphics[width=1.0\textwidth]{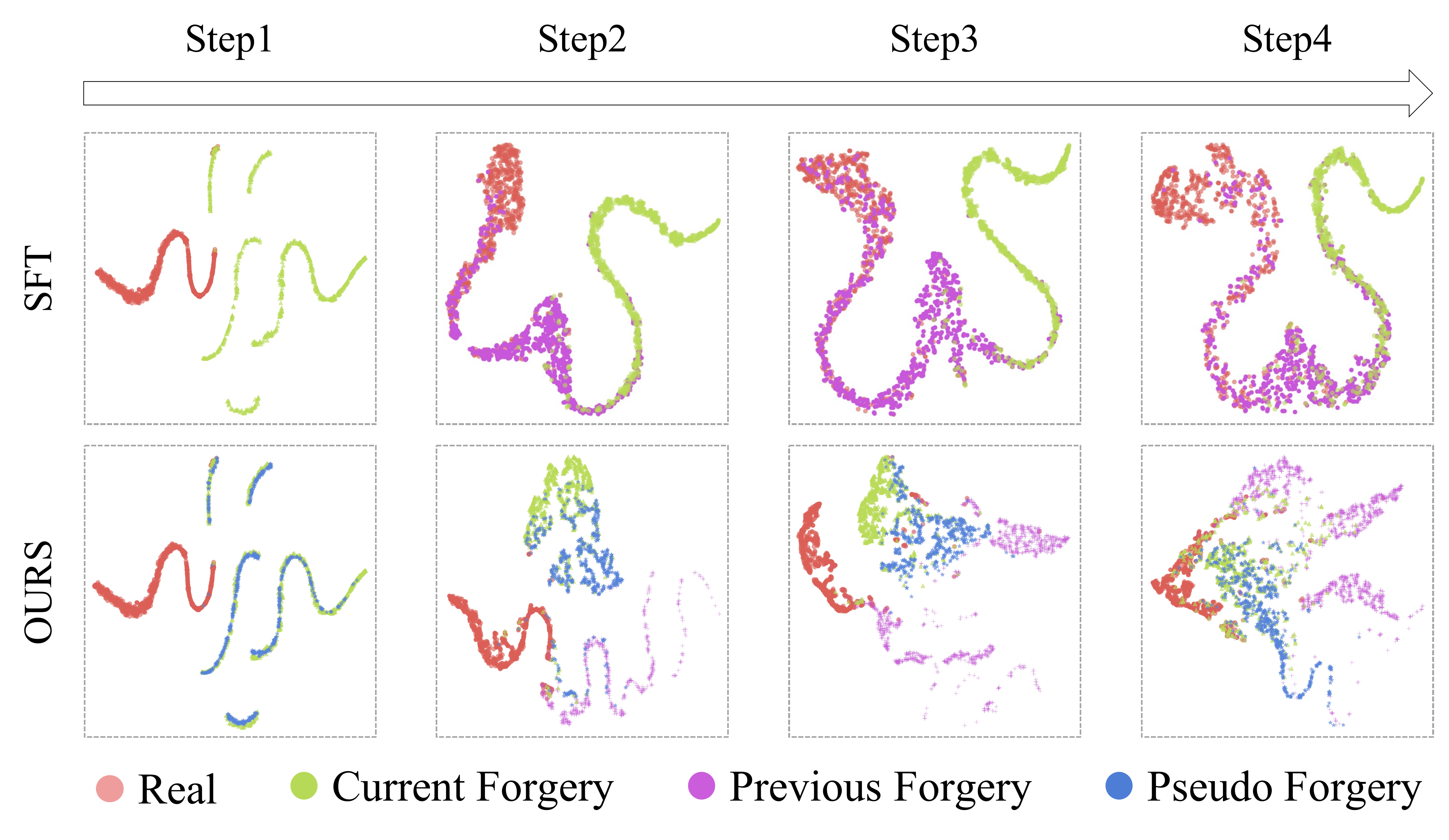}
    \end{center}
       \caption{
       Visualization of feature distribution on protocol1 via t-SNE. Step1 to step4 represents the Deepfakes, Face2Face, FaceSwap and NeuralTextures, respectively. 
       }
    \label{fig:tsne}
\end{figure*}

\noindent\textbf{Analysis of Distribution Change.}
Our method intends to alleviate the catastrophic forgetting by rehearsing the simulated forgery distribution via UAP-based pseudo samples. 
To show the dynamic process of CFFD, we draw the feature distribution of baseline (SFT) and our method using t-SNE. Specifically, we visualize the four-step models \textit{i.e.} Deepfakes, Face2Face, FaceSwap, and NeuralTextures of protocol1 with real, corresponding forgery, corresponding pseudo samples and previous forgery feature distributions, respectively. The results are shown in Fig.~\ref{fig:tsne}. We can observe that 1)
for the SFT model, the overlap between previous forgery features and real features becomes more and more obvious with the increase of the attack, which demonstrates the phenomenon of catastrophic forgetting.
2) For our method, there exists an obvious decision boundary between the previous forgery feature and the real feature in every stage, which demonstrates the effectiveness of the anti-forgetting of our method. 3) The distribution of pseudo-forged features and actual forgery features 
is resembling and overlapping with each other.
This phenomenon can support the motivation that uses these pseudo samples to simulate the forgery distribution. 

\section{Practicals of CFFD}
As AIGC advances, the variety of manipulated faces increases, making it impossible for the detection model to acquire all types of training data at once. This results in poor performance of the model when confronted with unfamiliar domain data, which severely constrains the deployment of the detection model. Although some methods have been proposed to enhance the generalization of the model, even the SOTA generalization models are still far from practical application. For instance, on the Celeb-DF dataset, the best generalization method achieves an AUC of 84\% under the cross-domain setting, while the intra-domain performance is 99\% using the same backbone. Therefore, in real-world applications, the detection model should be updated along with the emergence of new forgery methods. A naive way to update the model is to combine the newly collected data and historical data for training, but this has two drawbacks: first, it will consume a lot of computational and storage resources; second, it will cause security issues such as data leakage. Therefore, in this paper, we focus on the problem of continual face forgery detection (CFFD), which aims to enable the model to learn new data 
and avoid forgetting the previous knowledge simultaneously. We believe that the setting of CFFD will contribute to the deployment and promotion of face forgery detection models.


\section{Conclusion}
In this work, we focus on a challenging and practical setting, named Continual Face Forgery face detection (CFFD), which
enables efficient learning to deal with continually new attacks while mitigating catastrophic forgetting. Specifically, 
we propose a novel historical distribution preserving framework, which preserves the previous forgery distribution based on Universal Adversarial Perturbation (UAP), and
the knowledge distillation is further introduced to
maintain distribution variation of real faces across models over different periods. Correspondingly, we also build a new benchmark for CFFD with three different evaluation protocols. Extensive experiments on the novel benchmarks demonstrate that our method outperforms other SOTA competitors at a lower storage cost. The results affirm the potential of our approach as a robust and efficient solution in the ongoing battle against face forgery and underline the significance of continual learning in cybersecurity.


\section{Acknowledgments}
\label{sec:ack}
This work was supported by National Key R\&D Program of China (No.2022ZD0118202), the National Science Fund for Distinguished Young Scholars (No.62025603), the National Natural Science Foundation of China (No. U21B2037, No. U22B2051, No. 62176222, No. 62176223, No. 62176226, No. 62072386, No. 62072387, No. 62072389, No. 62002305 and No. 62272401), and the Natural Science Foundation of Fujian Province of China (No.2021J01002,  No.2022J06001).
\bibliography{sn-article}

\end{document}